\pdfoutput=1

\documentclass[11pt]{article}

\usepackage[]{EMNLP2023}

\usepackage{times}
\usepackage{latexsym}

\usepackage[T1]{fontenc}

\usepackage[utf8]{inputenc}

\usepackage{microtype}

\usepackage{inconsolata}

\usepackage{CJKutf8} 
\usepackage{graphicx}
\usepackage{float}
\usepackage{algorithmic}
\usepackage{algorithm}
\usepackage{array}
\usepackage[caption=false,font=normalsize,labelfont=sf,textfont=sf]{subfig}
\usepackage{multirow}
\usepackage{booktabs}
\usepackage{amsmath}
\usepackage{colortbl} 

\title{BianQue: Balancing the Questioning and Suggestion Ability of Health LLMs with Multi-turn Health Conversations Polished by ChatGPT}

\author{
    Yirong Chen\textsuperscript{1}, Zhenyu Wang\textsuperscript{1}, Huimin Zheng\textsuperscript{1}, Xiaofen Xing\textsuperscript{1}, Zhipei Xu\textsuperscript{1}, Kai Fang\textsuperscript{1},\\ 
    {\bf Sihang Li\textsuperscript{1}, Junhong Wang\textsuperscript{1}, Jieling Wu\textsuperscript{2}, Qi Liu\textsuperscript{3}, Xiangmin Xu\textsuperscript{3,4}\Thanks{ Corresponding author. Email: xmxu@scut.edu.cn}} \\
  \textsuperscript{1}School of EE., South China University of Technology, Guangzhou, China \\
  \textsuperscript{2}Department of Children’s Health Care, Guangdong Women and Children Hospital, China \\
  \textsuperscript{3}School of Future Technology, South China University of Technology, Guangzhou, China \\
  \textsuperscript{4}Pazhou Lab, Guangzhou, China \\
  \{\texttt{eeyirongchen,202030242072,202210182130}\}\texttt{@mail.scut.edu.cn}, 
  \texttt{xfxing@scut.edu.cn}, \\
  \{\texttt{eexzp,202030240160,202030241044,eewjh}\}\texttt{@mail.scut.edu.cn}, \\
  \texttt{jieling3861@163.com}, \{\texttt{drliuqi,xmxu}\}\texttt{@scut.edu.cn}\\
 }

\begin{document}
\maketitle
\begin{abstract}
Large language models (LLMs) have performed well in providing general and extensive health suggestions in single-turn conversations, exemplified by systems such as ChatGPT, ChatGLM, ChatDoctor, DoctorGLM, and etc. However, the limited information provided by users during single turn results in inadequate personalization and targeting of the generated suggestions, which requires users to independently select the useful part. It is mainly caused by the missing ability to engage in multi-turn questioning. In real-world medical consultations, doctors usually employ a series of iterative inquiries to comprehend the patient's condition thoroughly, enabling them to provide effective and personalized suggestions subsequently, which can be defined as chain of questioning (CoQ) for LLMs. To improve the CoQ of LLMs, we propose BianQue, a ChatGLM-based LLM finetuned with the self-constructed health conversation dataset BianQueCorpus that is consist of multiple turns of questioning and health suggestions polished by ChatGPT. Experimental results demonstrate that the proposed BianQue can simultaneously balance the capabilities of both questioning and health suggestions, which will help promote the research and application of LLMs in the field of proactive health.\footnote{\url{https://github.com/scutcyr/BianQue}}
\end{abstract}

\section{Introduction}
Recently, Large language models (LLMs), e.g. ChatGPT~\citep{chatgpt}, LLaMA~\citep{touvron2023llama}, ChatGLM~\citep{zeng2023glm-130b}, have been extensively applied in various fields. Through high-quality instruction fine-tuning and reinforcement learning based on human feedback (RLHF)~\citep{instructgpt}, LLMs already have possessed stunning language comprehension, generation, and knowledge reasoning abilities. Overall, users are amazed by the excellent suggestion ability of LLMs. 

\begin{figure*}[ht]
  \centering
  \includegraphics[width=\textwidth]{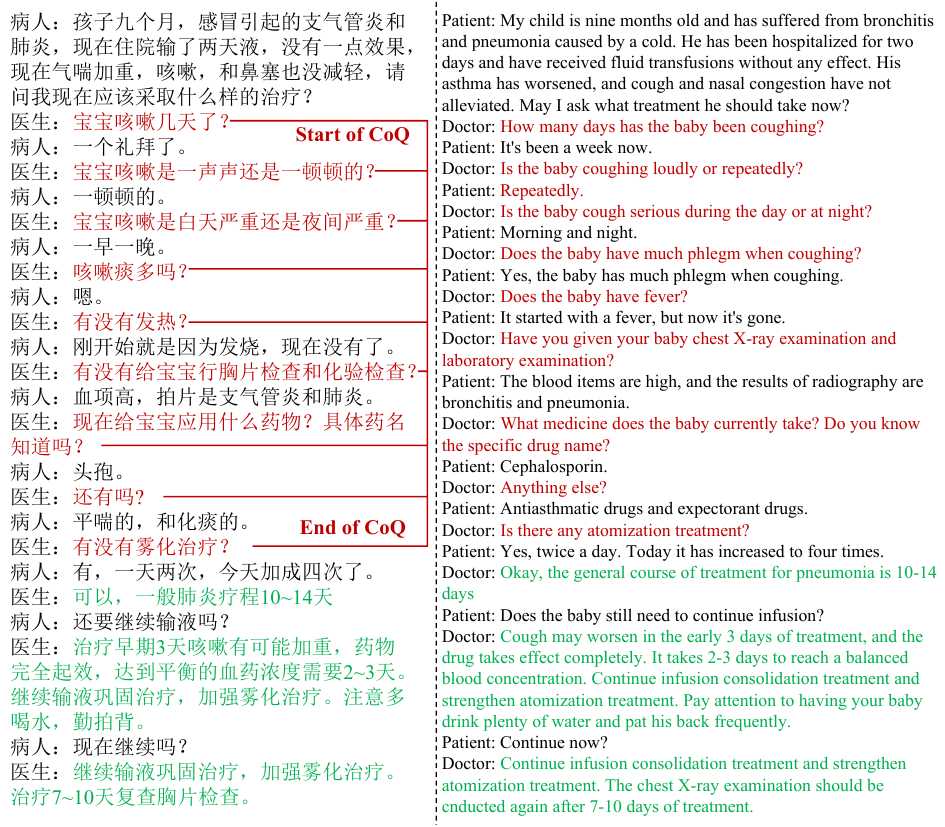}
  \caption{Example of chain of questioning (CoQ). The sentence in red font presents the doctor's CoQ: a series of questions about cough time, cough sound, sputum status, fever status, examination, medication and treatment.
}
  \label{CoQ_Example}
\end{figure*}

However, LLMs are deficient in ``questioning'' which is an important way to proactively understand users needs in medical, psychological, educational and other application scenarios. When we engage in healthcare conversations with these LLMs (ChatGPT\footnote{\url{https://chat.openai.com}}, ChatGLM\footnote{\url{https://chatglm.cn}}, SparkDesk\footnote{\url{https://xinghuo.xfyun.cn}}), they do not yet possess the ability to conduct multiple rounds of questioning, as presented in Appendix~\ref{sec:appendix_sample}. The above LLMs generally provide reasonable and universal suggestions based on the single-turn instruction provided by users. However, in the real world, doctors often need to conduct multiple turns of questioning with patients in order to provide targeted advice, as shown in Figure~\ref{CoQ_Example}. During the user consultation, the doctor raised different questions in the first 9 turns of conversations to understand the specific situation of the baby. The above multi-turn questioning process can be defined as \textbf{Chain of Questioning (CoQ)}. We found that the current LLMs lack CoQ capabilities because LLMs lack training data for multiple rounds of questioning during the instruction fine-tuning stage and RLHF stage. When researchers construct instructions and answers, on the one hand, they ignore multiple rounds of conversation history, and on the other hand, answers are usually suggestions rather than questions. 

At present, research on LLMs in the health field mainly focuses on evaluating the performance of existing models, constructing suitable datasets, and fine-tuning instructions. \citet{singhal2022large} proposed a medical Q\&A benchmark MultiMedQA for evaluating the clinical knowledge QA abilities of LLMs. \citet{li2023chatdoctor} constructed a real doctor-patient dialogue dataset HealthCareMagic-100k, and used it to fine-tune the ChatDoctor based on LLaMA. Similar health LLMs have been released one after another, e.g. BenTsao (\begin{CJK}{UTF8}{gbsn}本草\end{CJK})~\citep{wang2023huatuo}, ChatGLM-6B-Med~\citep{wang2023huatuo}, DoctorGLM~\citep{xiong2023doctorglm}, MedAlpaca~\citep{han2023medalpaca}, ClinicalGPT~\citep{wang2023clinicalgpt} and etc. These models are basically based on the assumption that ``users can clearly describe their problems or situations''. Therefore, during the model construction phase, \textbf{the questioning ability of the model was not considered}. Although these models have achieved well performance in the field of medical QA, they do not have the ability to ask users questions.

\begin{figure}[htbp]
  \centering
  \includegraphics[width=0.5\textwidth]{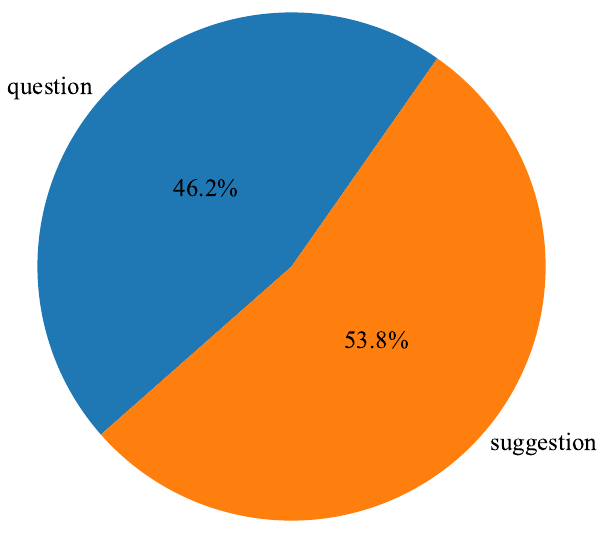}
  \caption{Proportion of questions and suggestions in answers of \textit{BianQueCorpus}.
}
  \label{answer_type}
\end{figure}

\begin{figure*}[ht]
  \centering
  \includegraphics[width=\textwidth]{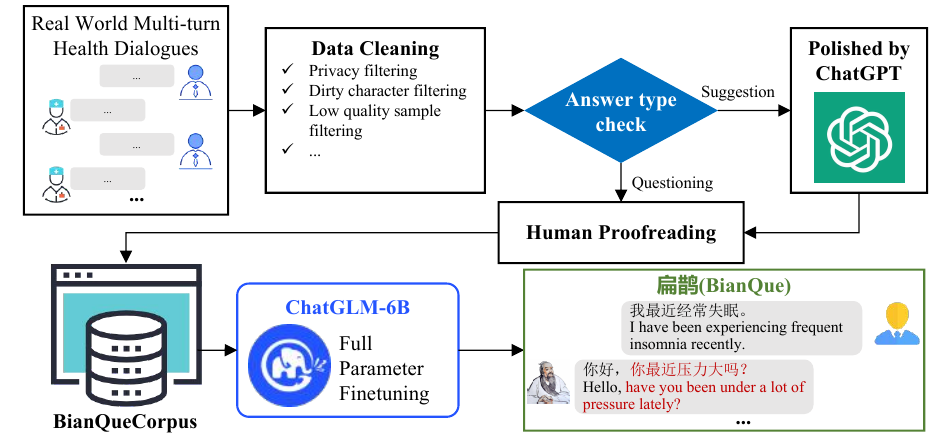}
  \caption{Construction process of \textit{BianQueCorpus} dataset and BianQue model.
}
  \label{BianQue_Construction}
\end{figure*}

To enhance the questioning ability of LLMs, we constructed a multi-turn health conversation dataset named \textit{BianQueCorpus}, in which the targets consist of balanced proportional questions ($46.2\%$) and suggestions ($53.8\%$), as shown in Figure~\ref{answer_type}. Meanwhile, we present BianQue, a health LLM that is specifically designed for balancing the questioning and suggestion ability. The results on multi-turn health conversation dataset demonstrate that BianQue outperforms existing models and ChatGPT, especially in the ability to conduct chain of questioning (CoQ).



\section{Methodology}
\label{sec:bianque}

\subsection{BianQueCorpus: Balancing Questioning and Suggestion}
\label{sec:bianquecorpus}
\begin{figure}[htbp]
  \centering
  \includegraphics[width=0.48\textwidth]{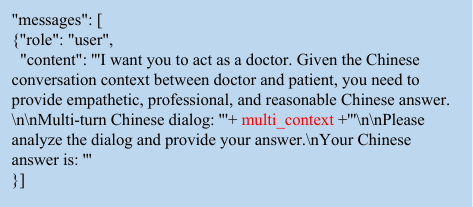}
  \caption{The prompt used for polishing the suggestions of doctors based on real-world multi-turn conversation context.
}
  \label{prompt}
\end{figure}

In the field of Chinese health conversational AI, there are already some multi-turn conversation datasets, e.g. MedDialog-CN~\citep{he2020meddialog}, IMCS-V2~\citep{10.1093/bioinformatics/btac817}, CHIP-MDCFNPC~\citep{zhang-etal-2022-cblue}, MedDG~\citep{zhang-etal-2022-cblue}. However, these conversations are often crawled from internet consultation platforms, such as \begin{CJK}{UTF8}{gbsn}好大夫\end{CJK}\footnote{\url{https://www.haodf.com/}}. These datasets are often mixed with a large amount of noise, such as missing content, missing images, reward information, privacy content, incomplete JSON content, website link, website tips, voice recording, text automatically replied by the system, etc. We first collected real-world multi-turn health conversations through data outsourcing services. Then, we performed a two-stage data optimization process: (i) We constructed a data automatic cleaning process based on regularized expression to improve the quality of existing conversation datasets. (ii) We designed a polishing prompt (see Figure~\ref{prompt}) and use ChatGPT to polish the doctors' suggestion of multi-turn conversations, because doctors often respond very briefly through internet platforms, lacking detailed analysis and suggestions. The whole construction process of \textit{BianQueCorpus} is presented in Figure~\ref{BianQue_Construction}. We ultimately obtained a multi-turn health conversation dataset consisting of 2,437,190 samples, in which the questions accounted for 46.2\% among the doctors' answers.

\subsection{BianQue Model}
We chose the ChatGLM-6B~\citep{du-etal-2022-glm, zeng2023glm-130b} as the base LLM architecture to construct the BianQue, since it is open source and has excellent Chinese understanding and generation performance. The input of model is defined as: 
$$
input = u_1^u+'{\backslash}n'+u_1^p+ ...+u_N^u+'{\backslash}n'+u_N^p
$$
where $u_i^u$=`\begin{CJK}{UTF8}{gbsn}病人：\end{CJK}' + utterance$_{i}^u$, $u_i^p$=`\begin{CJK}{UTF8}{gbsn}医生：\end{CJK}' + utterance$_{i}^p$ ($i<N$), $u_N^p$=`\begin{CJK}{UTF8}{gbsn}医生：\end{CJK}', $N$
is the number of dialogue turn.

\begin{table*}[htbp]
\caption{\label{evaluation_results} Evaluation results.}
\centering
\begin{tabular}
{clcccccccc}
\toprule
\small{Dataset} & \small{Model} & \small{BLEU-1} & \small{BLEU-2} & \small{BLEU-3} & \small{BLEU-4} & \small{R-1} & \small{R-2} & \small{R-L} & \small{PQA}\\
           

\midrule
\multirow{4}{*}{\small{\shortstack{\small{MedDialog} \\\small{-CN}}}} 
& \small{ChatGLM-6B} & \small{7.28} & \small{3.72} & \small{2.10} & \small{1.23} & \small{10.86} & \small{0.92} & \small{7.43} & \cellcolor{blue!20} \small{0.20} \\
& \small{DoctorGLM} & \small{10.39} & \small{5.06} & \small{2.94} & \small{1.80} & \small{13.27} & \small{1.04} & \small{11.17} & \cellcolor{blue!20} \small{0.01} \\
& \small{ChatGPT} & \small{7.61} & \small{3.90} & \small{2.21} & \small{1.30} & \small{11.11} & \small{0.96} & \small{7.82} & \cellcolor{blue!20} \small{0.28} \\
& \small{BianQue} & \small{\textbf{11.12}} & \small{\textbf{6.50}} & \small{\textbf{4.42}} & \small{\textbf{3.10}} & \small{\textbf{15.55}} & \small{\textbf{2.15}} & \small{\textbf{12.96}} & \cellcolor{blue!20} \small{\textbf{0.53}}\\

\midrule
\multirow{4}{*}{\small{IMCS-V2}} 
& \small{ChatGLM-6B} & \small{6.83} & \small{3.61} & \small{2.12} & \small{1.30} & \small{10.24} & \small{1.03} & \small{7.26} & \cellcolor{blue!20} \small{0.36} \\
& \small{DoctorGLM} & \small{8.38} & \small{4.22} & \small{2.52} & \small{1.55} & \small{11.87} & \small{0.95} & \small{9.22} & \cellcolor{blue!20} \small{0.06} \\
& \small{ChatGPT} & \small{8.46} & \small{4.54} & \small{2.71} & \small{1.70} & \small{11.48} & \small{1.29} & \small{8.97} & \cellcolor{blue!20} \small{0.38} \\
& \small{BianQue} & \small{\textbf{14.50}} & \small{\textbf{10.16}} & \small{\textbf{7.85}} & \small{\textbf{6.23}} & \small{\textbf{21.73}} & \small{\textbf{6.24}} & \small{\textbf{19.09}} & \cellcolor{blue!20} \small{\textbf{0.70}} \\

\midrule
\multirow{4}{*}{\small{\shortstack{\small{CHIP-} \\\small{MDCFNPC}}}}
& \small{ChatGLM-6B} & \small{6.22} & \small{3.11} & \small{1.81} & \small{1.10} & \small{9.62} & \small{0.85} & \small{0.67} & \cellcolor{blue!20} \small{0.35} \\
& \small{DoctorGLM} & \small{8.59} & \small{4.33} & \small{2.68} & \small{1.71} & \small{12.05} & \small{1.11} & \small{9.68} & \cellcolor{blue!20} \small{0.05} \\
& \small{ChatGPT} & \small{7.52} & \small{3.74} & \small{2.20} & \small{1.36} & \small{10.51} & \small{0.97} & \small{8.03} & \cellcolor{blue!20} \small{0.38} \\
& \small{BianQue} & \small{\textbf{13.41}} & \small{\textbf{8.49}} & \small{\textbf{6.05}} & \small{\textbf{4.42}} & \small{\textbf{19.00}} & \small{\textbf{3.99}} & \small{\textbf{16.56}} & \cellcolor{blue!20} \small{\textbf{0.57}}\\

\midrule
\multirow{4}{*}{\small{MedDG}} 
& \small{ChatGLM-6B} & \small{4.76} & \small{2.31} & \small{1.34} & \small{0.81} & \small{7.35} & \small{0.56} & \small{5.06} & \cellcolor{blue!20} \small{0.47} \\
& \small{DoctorGLM} & \small{6.87} & \small{3.47} & \small{2.15} & \small{1.35} & \small{9.62} & \small{0.88} & \small{7.61} & \cellcolor{blue!20} \small{0.09} \\
& \small{ChatGPT} & \small{5.11} & \small{2.41} & \small{1.38} & \small{0.83} & \small{7.58} & \small{0.50} & \small{5.46} & \cellcolor{blue!20} \small{0.63} \\
& \small{BianQue} & \small{\textbf{14.86}} & \small{\textbf{10.43}} & \small{\textbf{8.09}} & \small{\textbf{6.37}} & \small{\textbf{21.56}} & \small{\textbf{6.46}} & \small{\textbf{19.56}} & \cellcolor{blue!20} \small{\textbf{0.81}} \\

\bottomrule
\end{tabular}
\end{table*}

\section{Experiments}
\label{sec:experiments}

\subsection{Baselines and Benchmarks}
We select ChatGLM-6B\footnote{\url{https://github.com/THUDM/ChatGLM-6B}}~\citep{zeng2023glm-130b}, ChatGPT (gpt-3.5-turbo)~\citep{chatgpt}, and DoctorGLM~\citep{xiong2023doctorglm} as the baseline models. Comparative experiments were conducted on test set of MedDialog-CN, IMCS-V2, CHIP-MDCFNPC and MedDG respectively, since they are multi-turn conversation datasets that have both suggestions and questions in targets.

\subsection{Implementation details}
BianQue is finetuned on the proposed \textit{BianQueCorpus} using the \textit{WarmupDecayLR} learning rate scheduler with $warmup\_steps = 1000$ and $warmup\_max\_lr=5e-5$. During the training stage, the maximum input length is set to 1,536, while the maximum target length is set to 512. A batch size of 80 and global training steps of 25,000 are applied. The decoding algorithms of Top-p sampling with $p=0.75$ and temperature $\tau=0.95$ is applied in the inference stage.

\subsection{Results and Analysis}
Following the ClinicalGPT~\citep{wang2023clinicalgpt}, We evaluated BianQue and other models with the metrics: BLEU-1/2/3/4~\citep{papineni-etal-2002-bleu} and ROUGE-1/2/L~\citep{lin-2004-rouge}. In addition, we define a new metric to measure the model's Proactive Questioning ability (PQA):
\begin{equation*}
\begin{aligned}
&PQA = \frac{2P_{q}R_{q}}{P_{q}+R_{q}}, \\
&P_{q}=\frac{Q_{tp}}{Q_{tp}+Q_{t\overline{p}}}, R_{q}=\frac{Q_{tp}}{Q_{tp}+Q_{\overline{t}\overline{p}}},
\end{aligned}
\end{equation*}
where
$Q_{tp}$ is the number of samples with both target and prediction are question, $Q_{t\overline{p}}$ is the number of samples with question target and suggestion prediction, $Q_{\overline{t}\overline{p}}$ is the number of samples with both target and prediction are suggestion.

As shown in Table~\ref{evaluation_results}, BianQue demonstrates considerable performance on MedDialog-CN, IMCS-V2, CHIP-MDCFNPC and MedDG, achieving better scores than other models across all metrics.

\section{Conclusion and Future Work}
\label{sec:conclusion}
In this study, we introduced BianQue, a health LLM with balanced questioning and suggestion ability, which is finetuned based on the proposed large-scale multi-turn health conversation dataset \textit{BianQueCorpus}, in which the targets consist of balanced proportional questions ($46.2\%$) and suggestions ($53.8\%$). The empirical results highlight the superior mulit-turn questioning ability. Future work requires further focus on the conversion mechanism between questioning and suggestion.

\section*{Limitations}
It must be emphasized that there are potential risks when using generative language models for health conversation. Doctors in the real world are rigorous in diagnosing diseases and providing medication guidance. However, the current state-of-the-art LLMs (e.g. ChatGPT) still cannot guarantee the accuracy of the text they generate. Therefore, it is necessary to set up inspection and error correction mechanisms for the health suggestions generated by LLMs. At the same time, when LLMs learn the ability to proactively question, their usage risk also increases, as the models may ask users some questions related to privacy. For example, when users consult AI about cold-related issues, AI may proactively inquire about their age, gender, and other privacy information. Further privacy protection mechanisms need to be considered in the research and application of LLMs. Overall, the methods proposed in this article are still in the early research stage, and the questioning and suggestion mechanisms are not clear enough. The proposed model is limited to academic research and cannot be used in real-world deployment.

\section*{Ethics Statement}
The Bianque model is committed to improving the proactive questioning ability of LLMs, rather than providing very professional medical diagnosis or advice. The multi-turn conversation dataset used in this study is mainly based on the real world doctor-patient conversations, which has gone through a strict data cleansing process to eliminate private information and dirty text content. To this end, we constructed 50 regular expressions and used the \textit{re} package for filtering. We compared the data quality before and after data cleansing, and the excellent rate increased from 82\% to 93\%. Due to the lack of human feedback during the model finetuning stage, the current version of the model may involve user privacy when asking questions, which is particularly important to note. On the other hand, the health recommendations generated by the model have not undergone rigorous examination and proofreading, and therefore cannot be used as a substitute for real-world doctors. We emphasize that this is an early research-oriented model, rather than a mature and directly applicable model. Therefore, future work needs to combine RLHF to improve the safety level of model generated questions or suggestions. Besides, when Bianque is applied to downstream scenarios, it is necessary to inform the users in advance that the answers they see are generated by health AI, which are for reference only.

\section*{Acknowledgements}
This work was supported by the Science and Technology Project of Guangzhou (202103010002), the Natural Science Foundation of Guangdong Province (2022A1515011588), the National Key R\&D Program of China (2022YFB4500600), the Science and Technology Project of Guangdong (2022B0101010003), the National Natural Science Foundation of China under Grant U1801262 and Guangdong  Provincial Key Laboratory of  Human Digital Twin (2022B1212010004).



\appendix

\section{Reproducibility Checklist}
\label{sec:appendix_reproducibility}

\begin{itemize}
  \item
  \textbf{Model and Data:} The BianQue model and BianQueCorpus will be released upon decision of the paper.
  \item
  \textbf{System Hardware:} BianQue is trained on a Ubuntu 20.04.6 LTS server that has 2 CPUs called "Intel(R) Xeon(R) Platinum 8358P CPU @ 2.60GHz", 8 NVIDIA A800-SXM4-80GB GPUs, and 1,024GB memory.
  \item
  \textbf{Driver Version:} The version of Nvidia driver is "525.105.17". CUDA=11.6, and Cudnn=8.4.0.27.
  \item
  \textbf{Package version:} python=3.8.16, torch\footnote{\url{https://pytorch.org/get-started/previous-versions}}=1.13.1+cu116, transformers\footnote{\url{https://github.com/huggingface/transformers}}=4.28.0, deepspeed\footnote{\url{https://github.com/microsoft/DeepSpeed}}=0.9.3, datasets=2.11.0 and jieba=0.42.1 is recommended. Other dependent packages and versions will be released in our open source repository.
  \item
  \textbf{Model Parameters:} BianQue has 6.2B parameters with 28 layers and $max\_sequence\_length$ of 2,048. During the inference phase, the model requires at least 14GB of GPU memory.
  \item
  \textbf{Training Time:} BianQue is trained with global steps of 25,000 and $torch\_dtype$ of "float16" on 8 NVIDIA A800-SXM4-80GB GPUs. The training time is about 66 hours.

\end{itemize}

\section{Sample Conversations of LLMs}
\label{sec:appendix_sample}

\begin{figure*}[htbp]
  \centering
  \includegraphics[width=\textwidth]{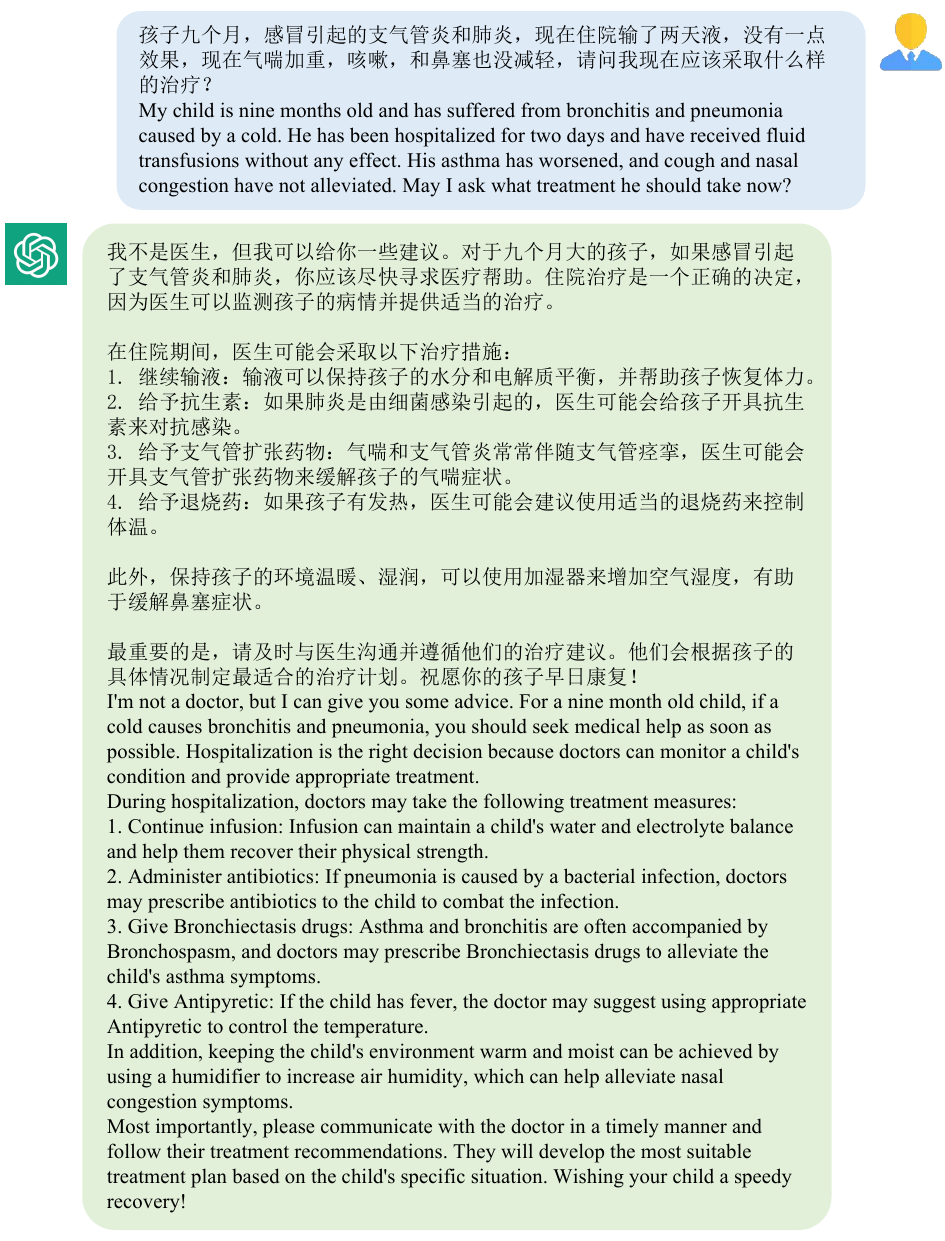}
  \caption{A case of a user confiding to ChatGPT.
}
  \label{ChatGPT_Example}
\end{figure*}

\begin{figure*}[htbp]
  \centering
  \includegraphics[width=\textwidth]{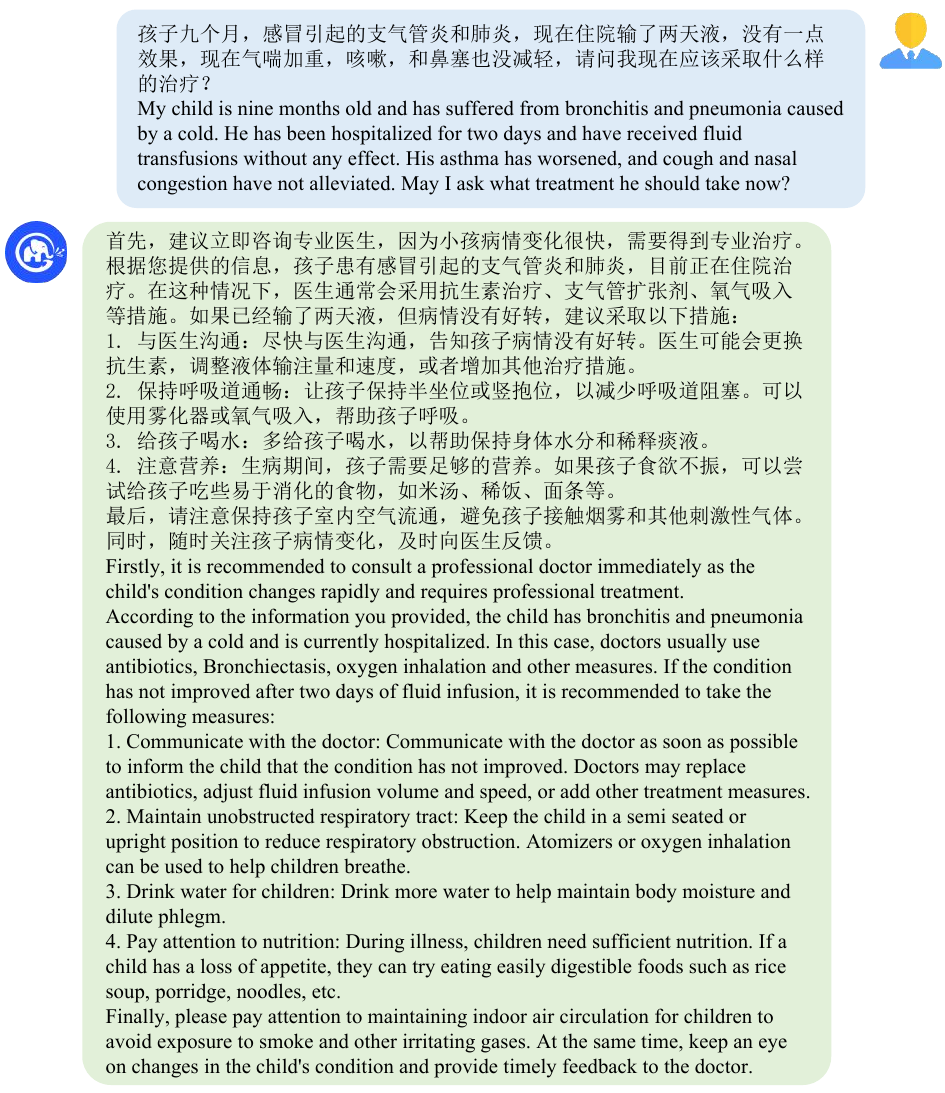}
  \caption{A case of a user confiding to ChatGLM.
}
  \label{ChatGLM_Example}
\end{figure*}

\begin{figure*}[ht]
  \centering
  \includegraphics[width=\textwidth]{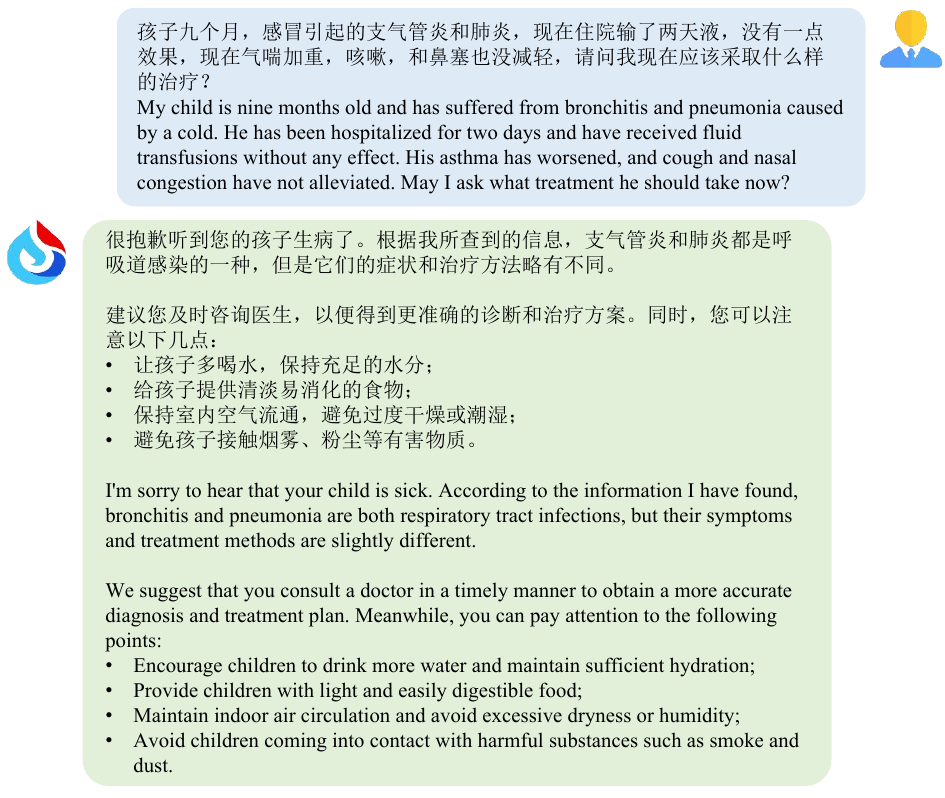}
  \caption{A case of a user confiding to SparkDesk.
}
  \label{SparkDesk_Example}
\end{figure*}

The following are examples of health conversation testing in ChatGPT (Figure~\ref{ChatGPT_Example}), ChatGLM (Figure~\ref{ChatGLM_Example}), and SparkDesk (Figure~\ref{SparkDesk_Example}). These are three common Chinese LLMs, but none of them have CoQ capabilities. The above LLMs generally provide reasonable and universal suggestions based on the single-turn instruction provided by users. However, in the real world, doctors often need to conduct multiple turns of questioning with patients in order to provide targeted advice.

\section{Sample Conversations of BianQue}
\label{sec:appendix_bianque}

\begin{figure*}[htbp]
  \centering
  \includegraphics[width=\textwidth]{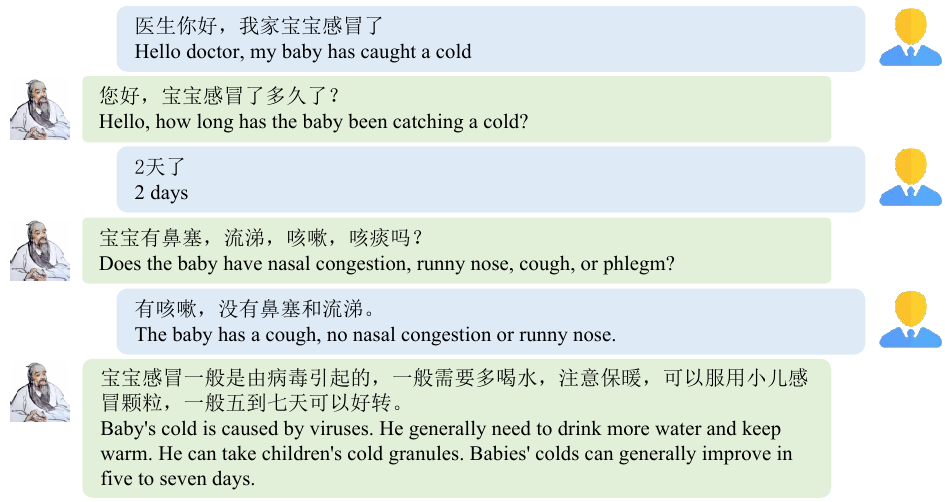}
  \caption{A case of a user confiding to BianQue.
}
  \label{BianQue_Example}
\end{figure*}

Figure~\ref{BianQue_Example} shows an example of the BianQue model proactively asking questions.

\end{document}